\pdfoutput=1

\documentclass[11pt]{article}

\usepackage[]{emnlp2022}

\usepackage{times}
\usepackage{latexsym}
\usepackage{comment}
\usepackage{svg}

\usepackage[T1]{fontenc}

\usepackage[utf8]{inputenc}

\usepackage{microtype}

\usepackage{cleveref}
\usepackage{makecell}
\usepackage{hyperref}
\usepackage{xurl}
\usepackage{graphicx}
\usepackage{xcolor}
\usepackage{subcaption}
\usepackage{tabularx}
\usepackage{booktabs}
\usepackage{multirow}
\usepackage{enumitem}
\usepackage{float}
\usepackage{wrapfig}

\title{The Esethu Framework: Reimagining Sustainable Dataset Governance and Curation for Low-Resource Languages}

\author{\normalsize Jenalea Rajab$^{1,3,*}$, Anuoluwapo Aremu$^{1*}$, Everlyn Asiko Chimoto$^{1*}$, Dale Dunbar$^{2,*}$, \\ 
\textbf{\normalsize Graham Morrissey$^{2,*}$, Fadel Thior$^{1}$, Luandrie Potgieter$^{1}$, Jessico Ojo$^{1}$, }\\
\textbf{\normalsize  Atnafu Lambebo Tonja$^{1,5}$, Maushami Chetty$^{6}$, Wilhelmina NdapewaOnyothi Nekoto$^{8}$, } \\
\textbf{\normalsize  Pelonomi Moiloa$^{1}$, Jade Abbott$^{1,7}$, Vukosi Marivate$^{1,4}$, Benjamin Rosman$^{1,3}$}\medskip \\
\footnotesize
$^1$ Lelapa AI, $^2$ Way With Words,  $^3$ RAIL Lab -- University of the Witwatersrand,  \\
\footnotesize
 $^4$ DSFSI -- University of Pretoria , $^5$ MBZUAI, $^6$ Aarya Legal, $^7$ Masakhane, $^8$Independent
\\
\footnotesize
$^*$ Equal Contributions
}

\begin{document}
\maketitle
\begin{abstract}

This paper presents the Esethu Framework, a sustainable data curation framework specifically designed to empower local communities and ensure equitable benefit-sharing from their linguistic resource. This framework is supported by the Esethu license, a novel community-centric data license. As a proof of concept, we introduce the Vuk’uzenzele isiXhosa Speech Dataset (ViXSD), an open-source corpus developed under the Esethu Framework and License. The dataset, containing read speech from native isiXhosa speakers enriched with demographic and linguistic metadata, demonstrates how community-driven licensing and curation principles can bridge resource gaps in automatic speech recognition (ASR) for African languages while safeguarding the interests of data creators. We describe the framework guiding dataset development, outline the Esethu license provisions, present the methodology for ViXSD, and present ASR experiments validating ViXSD’s usability in building and refining voice-driven applications for isiXhosa.
\end{abstract}

\section{Introduction}\label{sec:introduction}

The advancement of automatic speech recognition (ASR) systems has been underpinned by the availability of robust, high-quality speech data~\citep{10.1145/3491102.3517639,Shah2024SpeechRB,labelyourdata}. Over the past decade, an increasing number of such datasets have been released, spanning various languages, recording conditions and speaking styles ~\citep{solberg-ortiz-2022-norwegian,yang22h_interspeech,8099850,cífka2023jamaltformattingawarelyricstranscription,10.1162/tacl_a_00627}. Such diversity ensures that ASR systems are robust, fair, and effective for all members of a speech community. However, many low-resourced languages remain under served, leaving large communities without voice-driven technologies. In addition, data collection from these communities can often be exploitative, as current licenses do not ensure participants fairly benefit from the resources they help create~\citep{7439430,10.1145/3530190.3534792}. This underscores the need for more equitable and sustainable dataset creation and governance approaches.

\begin{figure}
    \centering
    \resizebox{\columnwidth}{!}{
        \includegraphics{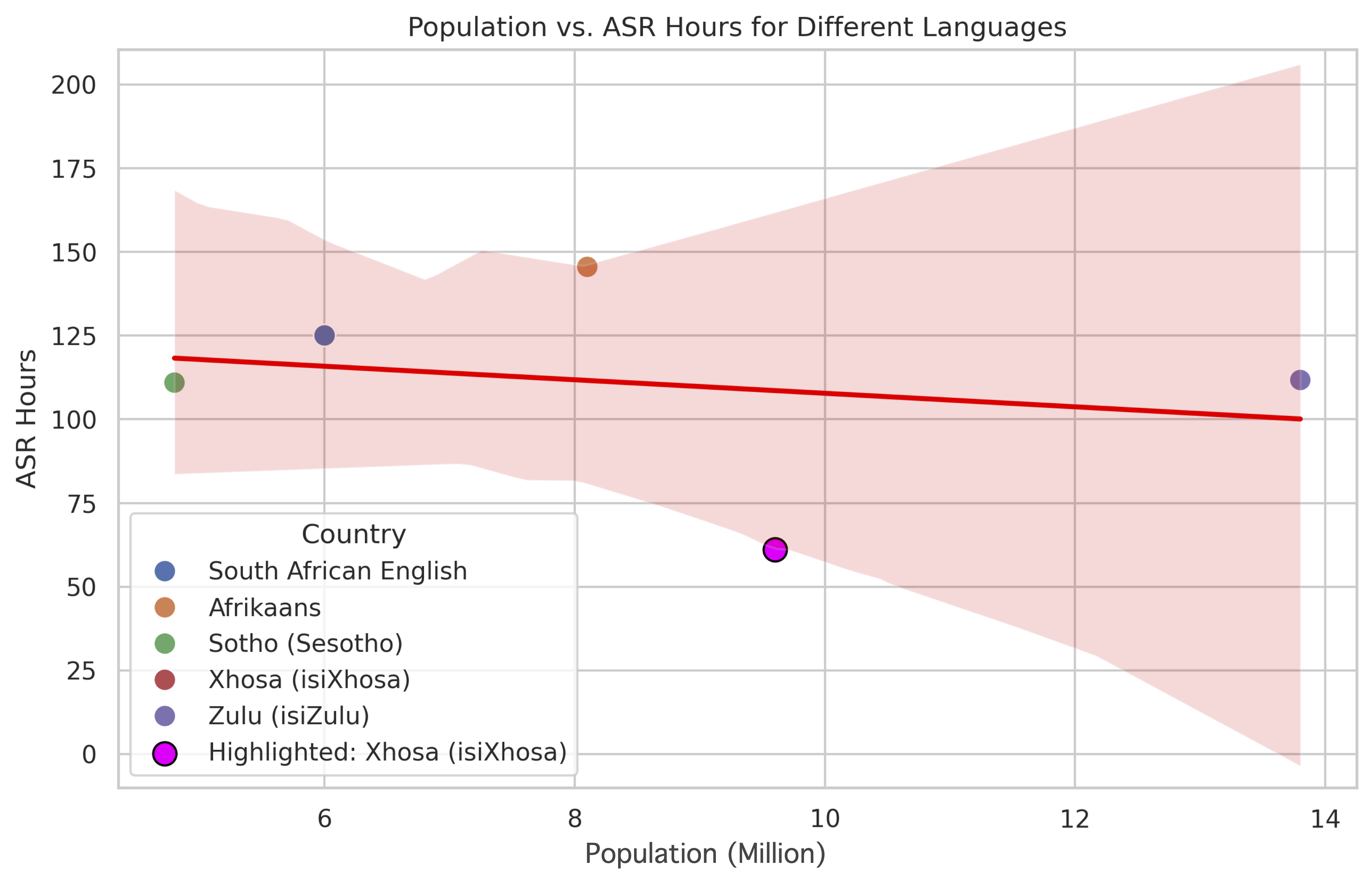}
        }
        \caption{Comparison between publicly available ASR data in hours for various African languages. isiXhosa falls below the trend line, indicating a shortage in publicly available ASR data.}
        \label{fig:setup}
\end{figure}

In this paper, we present the Esethu Framework, a pioneering data governance approach that fundamentally reimagines how low-resource language datasets are created, curated, and sustained. At its core, the framework introduces a novel economic model where licensing revenue is systematically reinvested into dataset expansion, ensuring continuous growth while directly benefiting the communities that create the data. This is supported by the Esethu License, a first-of-its-kind community-centric licensing scheme that maintains community agency over data assets while enabling broad research access and creating clear pathways for ethical commercialization. In particular, indigenous low-resourced African language speakers stand to benefit most immediately.

As a proof-of-concept, we develop and release the Vuk’uzenzele isiXhosa Speech Dataset (ViXSD). This dataset seeks to build inclusivity for the isiXhosa people and features diverse speakers, covering different genders, accents and backgrounds while demonstrating a crucial sustainable framework to ensure its growth. We detail the data creation process and framework as well as licensing with the aim of empowering researchers and developers, particularly in low-resource language settings, to create and refine ASR datasets and models.

We focus on isiXhosa, a Bantu language spoken by more than 9 million people in South Africa, as it epitomizes the low-resource gap~\cite{roux07_ssw}. As shown in \Cref{fig:setup}, isiXhosa is notably underserved, only providing approximately 61 hours of speech data despite its sizable population, which positions it below the comparative trend line for African languages. Moreover, existing datasets for under-resourced languages rarely offer continual economic benefits to the communities that supply the data. This situation highlights the importance of holistically addressing both data scarcity and community empowerment~\citep{le-ferrand-etal-2022-learning, nekoto2022participatory,nekoto-etal-2020-participatory}.

Our contributions can be summarised as as:
\begin{enumerate}
    \item We present a sustainable community-driven data curation framework that is geared for further creation of low-resource data and the empowerment of community members.
    \item We introduce a novel licensing scheme that allows communities to have agency over their data, ensuring they retain economic and decision-making benefits.
    \item We release an isiXhosa speech dataset, under the license in question, with extensive metadata of speakers and their backgrounds and include a validation experiment to ensure the data was effective for training.

\end{enumerate}

\section{Related Work}\label{sec:related}

The growing demand for speech datasets in underrepresented languages has driven various resource creation efforts. This section reviews notable  community-centered initiatives and African speech datasets, with a critical lens on their limitations regarding sustained community involvement and ethical licensing.

Multiple projects have aimed to address resource gaps in African languages through community involvement. \citet{nekotoparticipatory} introduced the Oshiwambo data set project, which created more than 5000 sentences in the Oshindonga dialect along with their English translations. This participatory and cost-effective approach highlighted the potential for collaborative resource creation in underrepresented African languages, but they were not
able to release the dataset due to the absence of an African-centered data license that would benefit the data creators \cite{african_licenses}. This highlights the persistent tension between open data dissemination and ethical stewardship of community-generated content. Similarly, AfroDigit presented the first audio digit dataset in 38 African languages enabling speech recognition applications such as telephone and street number recognition \citep{emezue2023afrodigits}. BibleTTS provides 86 hours of high-fidelity Bible recordings in 10 African languages~\citep{meyer2022bibletts}, and AfricanVoices focuses on collecting high-quality speech datasets for African languages while also providing speech synthesizers~\citep{ogun20241000}. AfriSpeech-200 extends these efforts by providing a 200-hour pan-African speech corpus for English-accented ASR in the clinical and general domains~ \citep{olatunji2023afrispeech}. Meanwhile, Kencorpus features 5.6 million words and 177 hours of speech of 3 Kenyan languages; Swahili, Dholuo, and Luhya~\citep{wanjawa2022kencorpus}. These datasets mark significant progress in resource availability, but remain top-down in structure, often lacking mechanisms for community governance and equitable attribution.

Several datasets have been developed to increase ASR resources specific to isiXhosa. \citet{louw2001african} collected a spontaneous monolingual corpus of isiXhosa and five other South African languages. This data, sourced from first-language speakers via an annotated telephone-based database. Although this work demonstrated the potential of African language resources, its scale and scope were limited. \citet{strom2018linguistic} captured linguistic diversity of isiXhosa by emphasizing the importance of including regional accents and tonal variations. Lastly, \citet{van2017synthesising} curated a multilingual dataset of four South African languages, including isiXhosa, in a code-mixed setting with English. This resource advanced acoustic modeling and multilingual ASR systems \cite{biswas2003semi}, demonstrating the potential of code-mixed datasets. However, its scope remained task-specific, limiting its utility for downstream applications.

In summary, while these initiatives demonstrate promising approaches to resource creation, they often fall short in embedding ethical, inclusive and sustainable models of community participation. Future work must prioritize not only linguistic coverage but also frameworks for ethical licensing, co-ownership, and long-term benefit sharing with the communities that contribute their languages.


\section{The Esethu Framework}
\label{sec:sustainableframework}
\begin{figure}[h!]
    \centering
    \resizebox{\columnwidth}{!}{
        \includegraphics{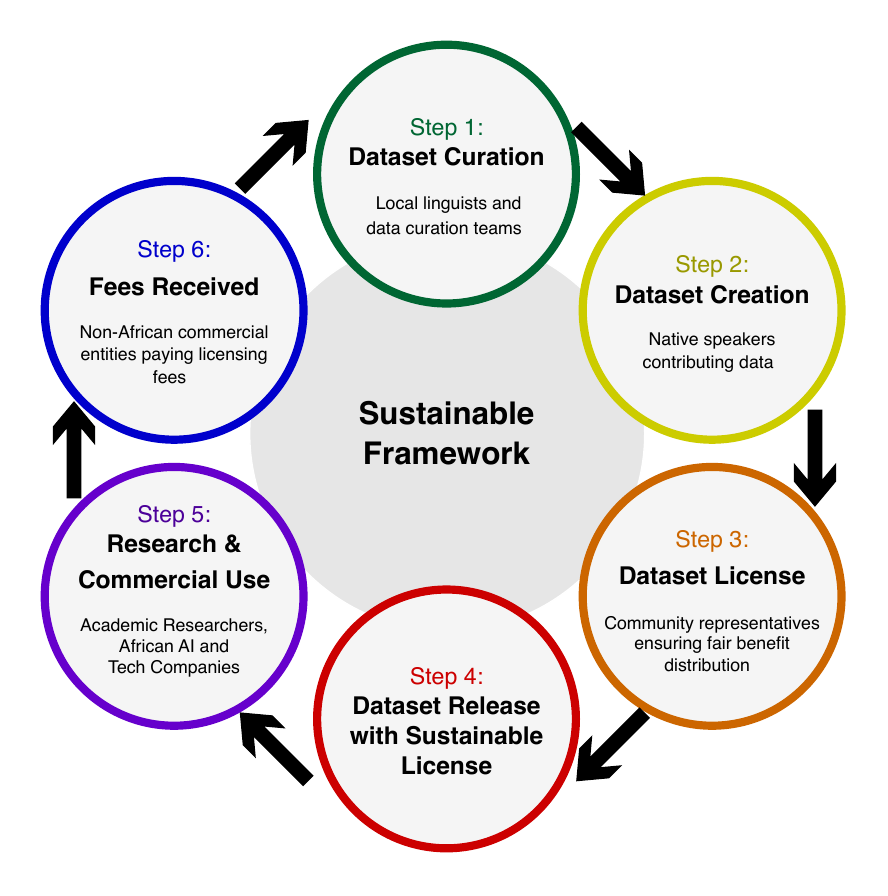}
        }
        \caption{Sustainable community-driven dataset curation framework.}
        \label{fig:framework}
\end{figure}

We define a community-driven dataset curation framework, illustrated in \Cref{fig:framework} to ensure reinvestment of language resources. This framework aims to address the challenges of limited data availability for low-resource languages by ensuring that the ownership and licensing of datasets remains with the native-speaking communities. The framework is supported by the development of a novel license, which aims to prevent external exploitation and ensures that speakers retain ownership over how their language is commercialized.
This decentralized control is particularly crucial for languages that have been historically marginalized, as it gives communities the agency to manage how their cultural and linguistic data is used,  fostering ownership and responsibility toward language preservation. 

As more communities adopt this framework, it enables the creation of a diverse set of datasets, ensuring that low-resource languages gain greater visibility and accessibility. Sustainability is achieved by reinvesting the fees associated with data licensing back into the community. This reinvestment strategy creates a self-sustaining ecosystem where financial resources fuel continuous improvement, data curation, and research initiatives.

\begin{figure}[t]
    \centering
    \includegraphics[width=\columnwidth]{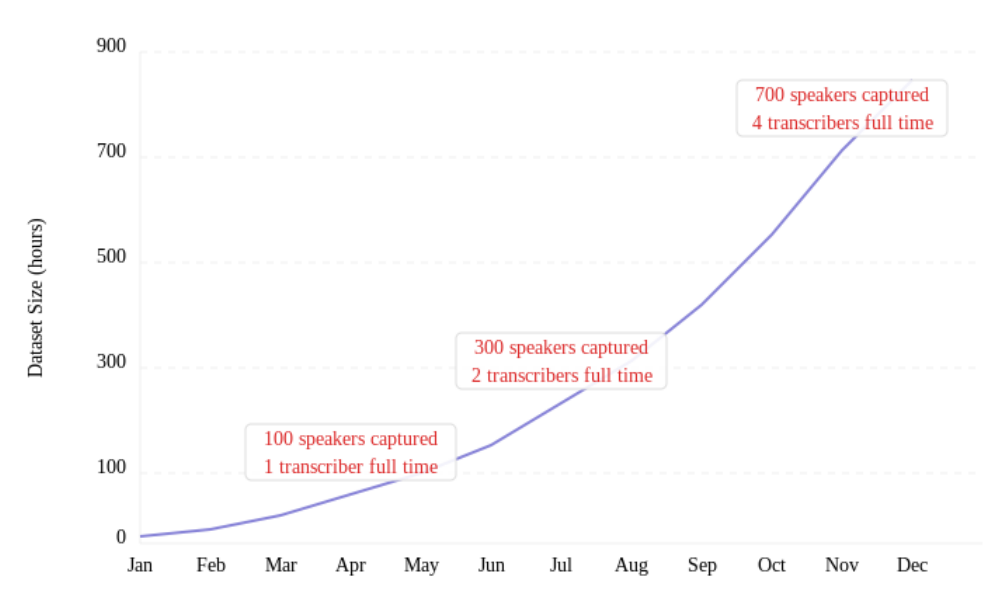}
    \caption{Dataset growth trajectory over twelve months, showing key milestones in speaker acquisition and transcriber employment. The growth is driven by a reinvestment model based on licensing revenue.}
    \label{fig:dataset-growth}
\end{figure}

To address the sustainability challenge in low-resource language dataset development, we provide a projection of the proposed reinvestment process that demonstrates scalability potential. Figure ~\ref{fig:dataset-growth} illustrates the dataset potential growth approach, where monthly licensing revenue—initialized at 1\% of initial dataset creation costs, set with a quarterly growth rate strategy of 20\%—is strategically reinvested into dataset expansion. This recursive investment mechanism yields substantial returns, potentially expanding the initial dataset from 10 hours to 893 hours over a twelve-month period. The efficacy is evident at three junctures: the initial milestone of 100 speakers (April), enabling the employment of one full-time transcriber; a subsequent expansion to 300 speakers (July) supporting two transcribers; and a final scale-up to 700 speakers facilitating a potential 4 full-time transcription positions. This progression not only demonstrates the process's potential financial viability but also its tangible impact on local language communities through sustained employment opportunities. The exponential growth trajectory (approximately 50-fold increase in dataset size) suggests that such a self-sustaining approach could significantly accelerate resource development for traditionally underserved languages while simultaneously creating stable employment within the target language community.

\subsection{Licensing} \label{sec:licensing}

Datasets are typically released under a range of licenses, chosen based on their intended use, applicable restrictions, and the goals of their creators. However, existing licensing frameworks do not adequately meet the unique requirements of this dataset, prompting the development of a custom licensing solution.

\textbf{Open and Permissive Licenses}: Licenses like the Apache and MIT Licenses are permissive but offer no protection against violations of external laws or third-party rights. In contrast, Creative Commons licenses impose specific conditions, such as the CC BY license, which requires attribution, and the CC0 license, which places data in the public domain. These licenses aim to promote open innovation by reducing legal barriers and have been successful in enabling both commercial and non-commercial use.

A significant limitation of these types of licenses lies in their underlying assumption that all users accessing the data possess relatively equal levels of agency. Although open access frameworks are intended to facilitate innovation by enabling access and reuse opportunities, the actual realisation of these opportunities is heavily dependent on user access to infrastructure, funding, and related resources. Furthermore, these licenses do not address the broader mechanisms of power that influence who can effectively utilize the data, thereby perpetuating inequalities despite their ostensibly inclusive design.

\textbf{Proprietary Licenses}: At the opposite end of the openness spectrum are closed proprietary datasets, which often require specific permissions or payments for use. Such licensing frameworks are typically used to serve particular motivations, including the need for control, monetization, or the protection of intellectual property. However, proprietary datasets are frequently criticized for their restrictive licensing terms and associated ethical concerns. These concerns include the potential exploitation of public data and violations of privacy, particularly given the lack of transparency regarding the contents of these datasets. The proprietary nature of such data prevents external validation, further exacerbating issues related to accountability and trust.

\begin{table*}[h]
    \centering
    \resizebox{\textwidth}{!}{%
    \begin{tabular}{lcccccc}
        \toprule
        \textbf{Language}  & \textbf{Speaker Population} & \textbf{Region} & \textbf{Dialects} & \textbf{Phonology} & \textbf{Morphology} & \textbf{Sounds}\\
        \midrule
        isiXhosa
        & $\sim$19M\footnote{\url{https://southafrica-info.com/arts-culture/11-languages-south-africa/}}
        & \makecell[l]{Eastern Cape ($\sim$62\%)\\%
                     Western Cape ($\sim$17\%)\\%
                     Gauteng ($\sim$9\%)}
        & 13\footnote{\url{https://glottolog.org/resource/languoid/id/xhos1239}}
        & \makecell[l]{10 vowels,\\58 consonants}
        & Agglutinative, 15 noun classes & \makecell[l]{Pulmonic egressive\footnote{such as those found in English}\\%
                     Velaric ingressive  (clicks)\\%
                     Glottic ingressive (implosive)}\\
        \bottomrule
    \end{tabular}%
    }
    \caption{Characteristics of the isiXhosa Language.}
    \label{tab:language}
\end{table*}

\textbf{Alternative Governance Models}: Community-owned corporations and co-operatives promote ethical commercialization of data, provided that reinvestment mandates exist in their founding documents. However, these models often restrict open access, limiting academic usage unless further provisions are made; or
require substantial legal, financial, and administrative overhead to set up and maintain, which may be prohibitive for many low-resource language communities. Beyond cooperatives, other governance models include data trusts, where data is held and managed by a trustee on behalf of a defined group of beneficiaries, in this case, language communities. While they offer a clear fiduciary duty to the community, establishing a data trust involves complex legal frameworks and may lack flexibility for dynamic participation over time.

\textbf{The Esethu License}: Datasets governed by both open and proprietary licenses are utilized by commercial entities for capital gain. However, it remains evident that those best positioned to capitalize on these datasets are often separated from those constrained by infrastructure and resource limitations, thereby reinforcing power imbalances and exacerbating global inequalities.

In the African context, access to these resources is severely limited, prompting the question of whether releasing African language datasets under traditional open or proprietary licenses genuinely supports the African language technology ecosystem. Releasing these datasets under open licenses exposes the African NLP ecosystem to external players–often non-African entities–that are better equipped with resources. Conversely, proprietary licensing may restrict the dissemination of opportunities within the African NLP ecosystem, especially if datasets are behind paywalls.

The dataset presented in this paper was developed with an ethical approach to data licensing as a central consideration in the release process. The creators of the dataset are committed to ensuring that the primary beneficiaries of the dataset are: (a) the contributors, in this case, the isiXhosa community, and (b) the broader African language technology ecosystem, for which no existing licensing have adequately met the needs. As a result, a new licensing model was devised to address these gaps.

Inspired by initiatives such as the Nwulite Obodo Data Licenses \cite{license} and the Kaitiakitanga Māori Data Sovereignty Licenses \cite{licence1}, we have developed a novel license for this dataset. This license consists of two components: a commercial license and an open license. In essence, the license does not restrict research use (inline with the Creative Commons BY-NC-SA license) and permits commercial use by African entities. We define African entities as those headquartered in Africa or majority-African-owned; case-by-case waivers permit diaspora or off-continent organizations with demonstrable ties. Non-African commercial entities are required to pay a licensing fee, see \Cref{AppendixA1}. We highlight legal strengths and weaknesses of Esethu license in \Cref{Appendix:licences}. 

Proceeds from the dataset are legally mandated to be reinvested in the creation of more data for the language in question, or less resourced languages, through a local data creation partner, see \Cref{AppendixA2}. This ensures a circular system that not only promotes the generation of additional data, benefiting the language community in question through language technology opportunities, but also guarantees that the revenue generated from the dataset supports fairly paid work for speakers. By ensuring that the data is accessible to African commercial entities only, we aim to foster innovation within the African language ecosystem while preserving opportunities for research on the language. 

\section{Vuk’uzenzele isiXhosa Speech Dataset
(ViXSD)}\label{sec:results} 
As a proof of concept, we develop and release the Vuk’uzenzele isiXhosa
Speech Dataset (ViXSD), developed with the Esethu framework in mind, and release it under the Esethu License with the help of a data curation partner. The data partner handles licensing administration, including collecting and allocating commercial fees. These fees would be reinvested into community-oriented initiatives, supporting continued data collection, compensating contributors, and expanding the dataset. This section describes the details the language itself, the dataset curation and creation.

\subsection{isiXhosa}\label{sec:isixhosa}
isiXhosa is classified within the Niger-Congo language family, specifically under the Nguni branch, which includes languages such as isiZulu, siSwati, and isiNdebele~\citep{LanguageFamilies}. The language is spoken predominantly in South Africa and parts of Zimbabwe and Lesotho. In South Africa alone, there are approximately 8M native speakers and 11M second-language speakers, primarily located in the Eastern Cape, Western Cape, Northern Cape, and Gauteng provinces \citep{10.1016/j.csl.2021.101262}.  isiXhosa is recognized as one of South Africa’s official languages and holds an institutional status, meaning it is used and sustained by educational, governmental, and media institutions \citep{ethnologue}. Notably, it ranks as the second most widely spoken Bantu language in the country (see \Cref{fig:setup}).

As a Bantu language, isiXhosa exhibits several distinctive linguistic features that make it both intriguing and valuable for speech analysis and technology. As shown in \Cref{tab:language}, it has 58 consonants including 18 click consonants and 10 vowels which are rarely used the Latin script \citep{10.1016/j.csl.2021.101262}. It is agglutinative and consists of 15 noun classes. It is a tonal language featuring two phonemic tones: high and low, as well as three categories of consonant sounds: pulmonic egressive sounds, velaric ingressive sounds, and one glottic ingressive sound~\citep{vanderstouwe_xhosa}. These characteristics not only enrich isiXhosa’s linguistic complexity but also present unique opportunities for insights in speech research.

\subsection{Vuk’uzenzele South African Multilingual Corpus}\label{sec:vukuzenzele}

\footnotetext[1]{\url{https://southafrica-info.com/arts-culture/11-languages-south-africa/}}
\footnotetext[2]{\url{https://glottolog.org/resource/languoid/id/xhos1239}}
\footnotetext[3]{such as those found in English}

The Vuk’uzenzele South African Multilingual Corpus~\cite{lastrucci2023preparing} aids as an initial case study for the Sustainable Data Curation Framework, offering an aligned multilingual text dataset derived from editions of the South African government magazine, Vuk’uzenzele. While this corpus provides a valuable linguistic resource for analysing South Africa’s multilingual landscape and the evolution of topical discourse across current events, politics, entertainment, and public affairs. 

Additionally, the Vuk’uzenzele dataset was selected for its engaging and accessible text, ensuring that participants could read fluently and with natural verbal expression. The readability of the content facilitated more coherent and expressive speech recordings, allowing participants to engage meaningfully with the material. This enhanced fluency and natural prosody, make the dataset particularly suitable for speech-based isiXhosa NLP applications, where intonation, rhythm, stress and expressive variation are critical for model performance.


\subsection{Data Pre-processing}\label{sec:preliminary}
Although the extracted Vuk’uzenzele dataset offers a rich linguistic resource, it required two rounds of pre-processing to ensure the text was structured, coherent, and suitable for speech recording.

First, text from the original articles was manually segmented into shorter, manageable units of 90 to 120 words, ensuring that participants could read and record comfortably, even on mobile devices. Beyond length considerations, additional formatting inconsistencies introduced by the automated extraction process such as misplaced headings, subheadings, page numbers, captions, irregular spacing, and formatting errors in bulleted lists were identified and corrected.

Each PDF package provided to participants included article metadata, such as the issue number, title, and author (where available). This structured approach ensured that the final dataset is clear, accessible, and optimised for efficient remote recording, particularly for participants in rural areas who relied solely on mobile devices. By refining the text presentation and optimising the workflow, the preprocessing stage significantly improved data quality and recording efficiency, facilitating a seamless transition into the recording phase.

\subsection{Participant Selection and Demographics} 

To be eligible for participation, applicants were required to be native isiXhosa speakers based in South Africa, with the ability to record themselves reading from a PDF text document. Although using a laptop or personal computer was preferred, individuals without such technology could still participate if they could produce recordings of sufficient audio quality through other means. This approach ensured that technological accessibility did not unduly limit the diversity of the participant pool.

Recruitment drew on two main sources: (1) existing networks of recording and transcription contractors and (2) an expression-of-interest form circulated to potential candidates. This approach enabled gathering of contact information from willing participants as well as prioritizing equal representation of male and female speakers. As a result, the dataset reflects a balanced range of speaker demographics ensuring diversity in the voice data.

Each participant received a unique speaker number linked to their file data. This dataset included comprehensive demographic metadata namely: self-reported age range, gender, education level, occupation, multilingualism level, current location, birthplace, location of formative years, duration of current residence, and any disclosed disabilities (speech-related or otherwise). Collecting these details provided insight into the linguistic and regional diversity of the participant pool.

Given the geographic concentration of participants, the dataset reflects the regional phonetic characteristics and accent variations commonly found among isiXhosa speakers from these areas, providing valuable insights into dialectal influences and pronunciation patterns.

\subsection{Recording Process and Quality Control}

A structured recording protocol was implemented to ensure consistency and high-quality speech data. This protocol included standardized guidelines on file format, equipment use, and submission procedures, aiming to preserve both technical quality and linguistic integrity.

\textbf{Recording Setup and Guidelines:} Participants were instructed to record audio in WAV format, with guidance provided on free recording software such as Audacity. Detailed instructions were given
on how to submit and recordings could be submitted either by email or via shared links.

\textbf{Equipment Variation:} Among the eight participants, one used a mobile phone, three relied on internal laptop microphones, three utilised headset microphones connected to laptops, and one alternated between a headset microphone and a laptop's internal microphone.  While this variation in equipment introduced some acoustic diversity, the structured recording guidelines helped ensure that the overall dataset maintained a baseline level of quality and consistency.
\newpage
\textbf{Quality Assurance Measures:} To ensure recording quality, participants were first required to submit sample recordings. These were evaluated for background noise, reading fluency, and pacing. Based on this assessment, individualised feedback and recommendations for improvement were provided via email. For instance, some participants defaulted to English when reading numbers. These participants were instructed to read numbers in isiXhosa, such that at least half of the dataset contains numerals expressed in isiXhosa rather than in English. This approach helps preserve linguistic features that may be diminishing in modern speech due to language shift and increasing code-switching trends. Also, participants were instructed to read each prompt at a comfortable pace in a quiet room. To ensure high-quality audio, participants were  permitted to re-record any prompt if they misread, hesitated, or paused unnaturally. We then ran an automated text–audio alignment tool and manually spot-checked all flagged utterances for major errors. Only those recordings that were clear, intelligible, and closely matched the target text were retained in the dataset. The full list of speaker guidelines is provided in Appendix \ref{Appendix:guidelines}.


\begin{table*}[t!]
    \centering
    \resizebox{\textwidth}{!}{%
    \begin{tabular}{l|ccccccc}
        \toprule
        \multirow{2}{*}{\textbf{Split}}  & \multirow{2}{*}{\textbf{Participants}} & \textbf{Gender}& \textbf{Age}& \textbf{Total Duration} & \textbf{Size} &  \textbf{Avg Duration} & \textbf{SNR }\\
        & & male/female &<29/<39/<49  & h:mm:ss & \#transcriptions & mm:ss per file & dB\\
        \midrule
        train & 4 &  2/2& 1/3/0 &7:47:01&  304& 01:32 &-0.01\\ 
        dev & 2 &  1/1&  0/2/0 &1:33:00 & 65&  01:26& 0.44\\ 
        test & 2 &  1/1&  0/1/1 &0:40:13 & 26&  01:33& -0.15\\ 
        \bottomrule
    \end{tabular}%
    }
    \caption{Characteristics of the isiXhosa ASR data.}
    \label{tab:data}
\end{table*}

\subsection{Dataset}
\textbf{Dataset Composition:} ViXSD\footnote{\url{https://huggingface.co/datasets/lelapa/Vukuzenzele_isiXhosa_Speech_Dataset_ViXSD}} consists of 395 stereo audio recordings and corresponding transcriptions derived from the Vuk’uzenzele South African Multilingual Corpus. It contains a total of 10 hours of narrated speech isiXhosa narrated by 8 speakers (4 male, 4 female) with approximately 39,000 words. We split the data into train, dev and test split for ease of use. The characteristics of each split are detailed in Table \ref{tab:data}.

\textbf{Demographic and Linguistic Distribution:} The audios were narrated by equal representation of male and females with education levels ranging from NQF Level 5 to Bachelors Degrees. The speakers are within the age range of 18-40 years, with 12.5\% being between 18–29 years, 75.5\% between 30–39 years and 12.5\% being 40 years. We focus primarily on isiXhosa with occasional code-switching where the script included non-isiXhosa terms (e.g., organizational names in English). We designed our selection process to reflect inclusivity within isiXhosa-speaking South African society. Where a person was born, has grown up and lives has a big impact on their accent as well as the vocabulary they will use when speaking their language. These were self-identified by the speakers in 3 categories detailed in Table \ref{tab:geographic_dist}.

\begin{table*}[t!]
\centering

\resizebox{\textwidth}{!}{%
\begin{tabular}{c|c|c|c}
\hline
\textbf{Category}  & \textbf{Eastern Cape (\%)} & \textbf{Western Cape (\%)} & \textbf{KwaZulu Natal (\%)} \\ \hline
Born in    & 75.0    & 12.5    & 12.5  \\
Spent Majority of Childhood in  & 50.0   & 25.0    & 12.5  \\
Currently Living in   & 62.5  & 37.5  & -   \\  \hline
\end{tabular}}
\caption{Geographic Distribution of Speakers.}
\label{tab:geographic_dist}
\end{table*}

\textbf{Data Quality:} Participants recorded the dataset in stereo using personal devices, typically in non-studio environments. This introduces variances including domestic sounds such as children, animals, and environmental noises. However, these real-world conditions can improve the robustness of the ASR system, helping it generalize better across diverse, noisy environments. While the recordings were done on single microphone devices, the platform used simulated stereo audio by duplicating the mono signal, resulting in a dual-mono format to ensure consistency. This approach does not introduce spatial depth or directionality, but it offers practical benefits: improved playback compatibility, cross-platform consistency, and the ability to later integrate true stereo recordings without restructuring the dataset.
 
\textbf{Data Features:} The dataset comprises original recordings in WAV format, with transcriptions provided in CSV format. Metadata associated with the recordings includes speaker ID, age range, gender, level of education, and location. The speech type consists of read or narrated speech, with topics spanning socio-economic, health, and community-related issues. 

\subsubsection{ASR Data Preprocessing} 
For ASR modelling, we preprocessed the audio files, converting them to mono channel and resampling them to a standard 16kHz. We then normalized the transcriptions by removing non-discriminative characters such as punctuation and special symbols that do not contribute to the model's understanding of the spoken language. These preprocessing steps enhance the clarity and usability of the data, setting a solid foundation for ASR training.

\subsection{Effectiveness for ASR}

To validate the usability of the dataset for ASR, we perform zero-shot testing and adapter fine-tuning on the open-source \textbf{Massively Multilingual Speech (MMS)} \cite{pratap2023mms} model. We chose MMS because it expands speech recognition and synthesis capabilities across over 1000 languages, and contains many low-resource or underrepresented languages, such as isiXhosa. Moreover, MMS makes use of adapter training where only a small fraction of model weights are updated. Adapter training is suitable for low resource settings as it is memory efficient \citep{pratap2023mms} and has shown to yield better performance \citep{mainzinger-levow-2024-fine}, making it an ideal framework for evaluating and enhancing isiXhosa ASR capabilities with our newly created speech dataset.


\textbf{Training and testing Details:} We performed zero-shot testing on two of the MMS checkpoints (\texttt{mms-1b-fl102} and \texttt{mms-1b-all}) fine-tuned for speech recognition with 102, and 1162 adapter weights respectively (one for each language). We also fine-tune the \texttt{mms-1b-fl102} for 5, 10 and 15 epochs on the training set respectively and make these models available for public use\footnote{\url{https://huggingface.co/lelapa}}. Hyper-parameters were kept consistent across all training to ensure comparability of results. The batch size is set at 2 with gradient accumulation steps set at 16, giving a total train batch size of 32, and the learning rate is kept at 0.001.

\subsubsection{Results}

\begin{table}[h!]
    \centering
   \resizebox{\columnwidth}{!}{%
    \begin{tabular}{lll}
    \toprule
        \textbf{Model(s)}  &\textbf{WER} & \textbf{CER}\\
    \midrule
      \textit{Zero-Shot Testing}& &\\
    \midrule
        \texttt{facebook/mms-1b-fl102} & 0.356  & 0.066\\
        \texttt{facebook/mms-1b-all} &  0.372 & 0.068\\
    \multicolumn{3}{l}{\textit{Massively Multilingual Transfer}}\\
    \midrule
        \texttt{mms-1b-fl102-xho-5}   & 0.335 & 0.058\\
        \texttt{mms-1b-fl102-xho-10}  & 0.310 & 0.052\\
        \texttt{mms-1b-fl102-xho-15}  & 0.321 & 0.052\\
        
    \bottomrule
   \end{tabular}
    }
    \caption{Word Error Rate (WER) and Character Error Rate (CER) of the above models on the ViXSD test data.}
    \label{tab:NMT_results}
\end{table}


\begin{figure}
    \centering
    \includegraphics[width=1\linewidth]{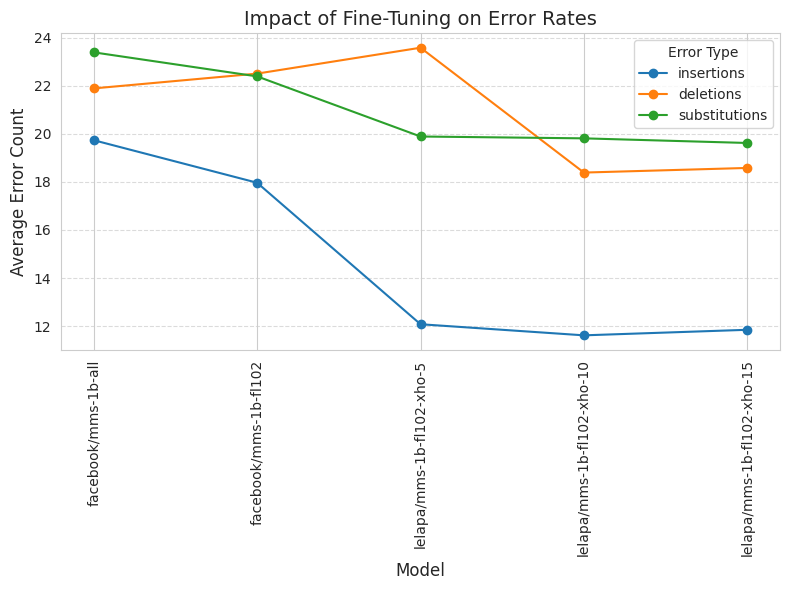}
    \caption{Image shows the reduction in common ASR errors when finetuned on the ViXSD data.}
    \label{fig:errors}
\end{figure}

Fine-tuning with the ViSXD data consistently improved ASR performance relative to the zero-shot baselines, as shown in \Cref{tab:NMT_results}. Across the epochs tested, all fine-tuned models outperformed the best baseline model (\texttt{mms-1b-fl102}), reducing the Word Error Rate (WER) by approximately 3.4\%. Notably, the optimal result (31\% WER) was achieved after ten epochs of fine-tuning, representing a 4.6\% improvement compared to the baseline’s 35.6\% WER. 

This unanimous decrease in WER underlines the effectiveness of combining transfer learning with the newly created ViSXD dataset. Additionally, the \texttt{mms-1b-fl102} model, which uses fewer compute resources, demonstrated a superior zero-shot WER (35.6\%) than the alternative baseline (\texttt{mms-1b-all}) (37.2\%), indicating that smaller-scale models can still achieve competitive performance while benefiting from focused fine-tuning in low-resource contexts.

To further evaluate the ASR transcription errors, we used a paired t-test to compare predictions from the zero-shot baselines to those of the fine-tuned models, shown in \Cref{fig:errors}. The results show a significant reduction in insertions (t = 10.27, p = 0.00197) and substitutions (t = 8.06, p = 0.00399), indicating that fine-tuning with the ViSXD dataset improves ASR performance and reduces hallucinations. However, the reduction in deletions (t = 0.91, p = 0.4288) was not significant, suggesting that fine-tuning has less impact on this error type.


\subsection{Lessons Learned}

\textbf{The Data Licensing Framework:} In defining the license, we encountered the challenge of identifying which enterprises qualify as “African” for fee exemption. In order to ensure inclusivity, we used two yardsticks. (a) businesses/companies registered and has its headquarter in an African country and (b) businesses/companies that are operational out-of-Africa but whose largest shareholding is African. This allows for the interest of Africans, regardless of their physical location, to be a beneficiary of this initiative.

\textbf{Transcriptions:} Transcripts should be long form and conversational. This facilitates both easy quality control and a smoother workflow. Technical text or acronyms hinder the transcription processes as annotators often need to consult with the project managers to standardize how these are spoken.

\textbf{Standardisation of Numerals:} A recurring issue we encountered was participants reading numbers in English, while others using isiXhosa. This introduces linguistic inconsistency and code-switching artifacts. To ensure consistency, we recommend setting numerical conventions upfront.

\textbf{Recording Equipment Considerations:} Participants employed various devices resulting in audio captured at different sample rates and stereo formats. While these differences required resampling and channel conversion, we observed no substantial quality gap between recordings. This suggests that strict device requirement may be unnecessary if proper recording guidelines are followed.

\section{Conclusion}\label{sec:discussion}

In this paper, we present the Esethu Framework, a pioneering approach to sustainable dataset governance that fundamentally reimagines how low-resource language data is created, licensed, and grown. Through its innovative reinvestment model—where licensing revenue is systematically channeled back into dataset expansion—the framework establishes a self-sustaining ecosystem for continuous resource development. As proof of concept, we developed the Vuk'uzenzele isiXhosa Speech Dataset (ViXSD), which demonstrates both the framework's practical viability and its potential impact: our experiments show that fine-tuning yields up to 4.6\% improvement in ASR performance, validating the technical quality of datasets produced under this approach. The Esethu Framework, with its emphasis on community ownership, sustainable growth, and ethical commercialization, offers a reproducible model for addressing the critical resource gaps in low-resource languages, ensuring that language communities remain primary beneficiaries of their data contributions while creating pathways for economic empowerment through language technology development.

\section{Limitations}

Given the ViXSD dataset focuses on news type articles, it may exhibit thematic and linguistic biases. Efforts have been made to preserve dialectal diversity and cultural authenticity, yet idiomatic expressions and culturally specific references may require careful interpretation in downstream NLP applications. Researchers should be cognisant of potential stereotypical or skewed linguistic representations when applying models trained on this corpus. Additionally, although efforts were made to ensure diverse speakers, the current dataset as it stands may not fully representative of the full socio-economic and geographical range of isiXhosa speakers. We also acknowledge that our current dataset remains relatively small, which may lead to models overfitting on specific speaker characteristics or acoustic environments. Future expansions using our proposed framework will include speakers from multiple provinces and regions to broaden coverage.

\section{Ethical considerations}

The creation of this ViXSD dataset necessitated careful consideration of ethical issues, particularly regarding data privacy, cultural representation, and linguistic bias. The dataset is derived from publicly available material, ensuring compliance with copyright and data protection regulations. Personally identifiable information (PII) is either absent or anonymised to safeguard individual privacy. Additionally, through the Esethu license, the data curators remain stewards of the dataset and any licensing fees for the data are reinvested into the development of the isiXhosa language speaker community, further data rights are detailed in Appendix \ref{Appendix:dataprotection}. The isiXhosa contributors were compensated at a fair market rate for their participation.

\section*{Acknowledgements}
We would like to thank the isiXhosa community speakers for contributing their voices, language, culture and heritage to this work.

\bibliography{anthology_0,anthology_1,custom}
\bibliographystyle{acl_natbib}

\clearpage
\appendix

\onecolumn
\section{Appendix}
\label{Appendix}

\subsection{Extracts from the Esethu License}
\label{AppendixA1}
1. We license the dataset for research and non-commercial purposes under the following license: \url{https://creativecommons.org/licenses/by-nc-sa/4.0/?ref=chooser-v1}
\\
\\
2. In summary:
\begin{enumerate}
    \item You may share – copy and redistribute the material in any medium or format;
\item You may adapt – remix, transform and build upon the material;
\item You must give appropriate credit, provide a link to the license, and indicate any
changes made;
\item You may not use the dataset for commercial purposes;
\item If you remix, transform, or build upon the material, you must distribute your
contributions under the same license as the original;
\item You may not apply legal terms or technological measures that legally restrict
others from doing anything the license permits.
\end{enumerate}

\noindent
3. We may permit commercial use of the dataset on the following terms:
\begin{enumerate}
\item Under a separate commercial license;
\item You must request access for commercial purposes by the commercial agreement with the data partner;
\item Commercial access shall be at a fee which fee shall be waived for African
entities;
\end{enumerate}

\subsection{Strengths and Weaknesses of the Esethu license}
\label{Appendix:licences}
Table \ref{tab:licences} provides an in-depth legal comparison of popular licenses with a focus on those developed for low-resource languages in the African context.

\subsection{Extracts from the Commercial Agreement}
\label{AppendixA2}

OWNERSHIP OF THE DATASET
\\
2.6 The Parties agree that, as at Signature Date, there is a dearth of isiXhosa and other African language datasets available for research and community purposes and, accordingly, anticipate that they may jointly create further datasets specifically intended for such purpose/s in the future (“Future Community Datasets”) and, in the event that they do so,  each party undertakes that it will at that time take the necessary steps in the creation of any such Future Community Dataset to ensure that same shall be original, that the Parties’ contributions thereto shall be indivisible and not ancillary to each other.
\\
\\
SHARING OF THE PROCEEDS
\\
5.1 The Parties hereby agree that, with effect from the Signature Date, any and all proceeds derived, from time to time, from the commercial use of the Community Dataset and any Future Community Datasets shall be paid to Party A for use as set out in clause 5.2.
\\
5.2 All proceeds from the commercialisation of the Community Dataset and/or Future Community Datasets shall be used solely for the public interest purpose of producing further IsiXhosa, or, as may be agreed by the Parties, other African language, Future Community Datasets.

\begin{table*}[]
\centering
\resizebox{\textwidth}{!}{
\begin{tabular}{p{2.5cm}p{3cm}p{2.5cm}p{2.2cm}p{3cm}p{4cm}}
\textbf{Metric}                       & \textbf{Nwulite Obodo}                                                 & \textbf{CC BY-SA 4.0}                     & \textbf{Apache 2.0}                    & \textbf{Kaitiakitanga}                      & \textbf{Esethu License}                                                                                 \\ \toprule
\textit{Privacy}                      & No explicit protections.                                                       & No provisions.                            & No provisions.                         & Implicit: Permission-based access.                  & No explicit protections.                                                                                                        \\ \hline 
\textit{Agency for Data Creators}     & High: Tiered control by community.                                             & Low: Licensor retains attribution rights. & Moderate: Broad permissions granted.   & Highest: Māori custodians retain veto power.        & High: Co-owners control commercialisation.                                                                                      \\ \hline
\textit{Socioeconomic Benefits}       & Royalties from non-developing countries can fund local data but not mandatory. & None.                                     & None.                                  & Conditional: Commercial terms negotiable.           & Explicit: Commercial fees fund future African language datasets (clause 5.2).                                                   \\ \hline
\textit{Attribution}                  & Required.                                                                      & Required.                                 & Required (NOTICE file).                & Unspecified.                                        & Required under CC BY-SA for non-commercial use; commercial terms require attribution to data owners “where possible”.           \\ \hline
\textit{Non-Commercial Use}           & Free for Developing Countries; fees for others.                                & Allowed.                                  & Allowed.                               & Default: Prohibited unless permitted.               & Dual: Non-commercial = CC BY-SA; commercial = Esethu (fee-based).                                                               \\ \hline
\textit{No Modifications/Derivatives} & Modifications allowed (ShareAlike).                                            & Modifications allowed (ShareAlike).       & Modifications allowed (no ShareAlike). & Restrictive: Derivatives bound by original license. & Non-commercial = CC BY-SA (ShareAlike); commercial = Esethu (negotiated).                                                       \\ \hline
\textit{ShareAlike}                   & Required for derivatives.                                                      & Required for derivatives.                 & Not required.                          & Strict: All derivatives bound by original license.  & Non-commercial = CC BY-SA (ShareAlike); commercial = Esethu (no ShareAlike).                                                    \\ \hline
\textit{Royalties}                    & Yes (non-developing countries). May include non-monetary benefit/interest.     & No.                                       & No.                                    & Potential (negotiated).                             & Yes: Commercial fees paid to partner for reinvestment.                                                                          \\ \hline
\textit{Open/Restrictive}             & Hybrid: Geofenced openness.                                                    & Fully Open.                               & Permissive.                            & Restrictive.                                        & Hybrid: Non-commercial = open (CC BY-SA); commercial = restrictive (Esethu).                                                    \\ \hline
\textit{Geographic Scope}             & Tiered (Developing vs. non-Developing).                                        & Universal.                                & Universal.                             & Unspecified (prioritizes Indigenous users).         & Universal: Commercial terms apply globally, but fees waived for African entities (Annexure A, term 3).                          \\ \hline
\textit{Data Sovereignty}             & Strong: Community reinvestment.                                                & Weak.                                     & Weak.                                  & Strongest: Embeds Indigenous governance.            & Strong: Co-owners retain control; proceeds fund African language datasets.                                                      \\ \hline
\textit{Enforcement of Derivatives}   & Prohibits export to non-Developing Countries.                                  & Global sharing allowed (ShareAlike).      & No restrictions.                       & Absolute: Derivatives must comply.                  & Non-commercial = CC BY-SA terms; commercial = negotiated.                                                                       \\ \hline
\textit{Termination}                  & Automatic if country reclassified.                                             & Violations only.                          & Patent litigation only.                & Implicit (violations = revoked access).             & Breach triggers termination (clause 8).                                                                                         \\ \hline
\textit{Patent Rights}                & Not addressed.                                                                 & Not addressed.                            & Explicit grant.                        & Not addressed.                                      & Not addressed.                                                                                                                  \\ \hline
\textit{Commercialization Control}    & Permitted but non-developing nations subject to a royalty.                     & No restrictions.                          & No restrictions.                       & Total control: Requires negotiation.                & Controlled: Commercial use requires fee (waived for African entities based on broad definition and case-by-case consideration). \\ \hline
\textit{License Compatibility}        & Compatible with similar community licenses.                                    & Compatible with CC BY-SA.                 & Compatible with most licenses.         & Incompatible: No relicensing allowed.               & Dual: CC BY-SA (non-commercial) + Esethu (commercial). \\ \bottomrule                                                                        
\end{tabular}}
\caption{Legal comparison of popular licenses with a focus on those developed for low-resource languages/the African context. We have used the metrics in the first column to compare them.}
\label{tab:licences}
\end{table*}

\newpage
\subsection{Speaker Guidelines}
\label{Appendix:guidelines}

\begin{enumerate}
    \item \textbf{Recording Format}
    \begin{enumerate}
        \item Record audio in WAV format.
        \item Use free recording software (e.g., Audacity) as suggested.
    \end{enumerate}
    
    \item \textbf{Equipment Use}
    \begin{enumerate}
        \item Use available recording devices: mobile phones, internal laptop microphones, or headset microphones.
    \end{enumerate}
    
    \item \textbf{Submission Process}
    \begin{enumerate}
        \item Submit recordings via email or shared link as instructed.
    \end{enumerate}
    
    \item \textbf{Quality Assurance}
    \begin{enumerate}
        \item Submit a sample recording for initial evaluation.
        \item Feedback will be provided on background noise, reading fluency, and pacing.
        \item Re-record prompts if there are misreads, hesitations, or unnatural pauses.
    \end{enumerate}
    
    \item \textbf{Language Use}
    \begin{enumerate}
        \item Read numbers in isiXhosa rather than English.
    \end{enumerate}
    
    \item \textbf{Reading Guidelines}
    \begin{enumerate}
        \item Read each prompt at a comfortable pace.
        \item Record in a quiet room to avoid background noise.
    \end{enumerate}
    
    \item \textbf{Post-Processing}
    \begin{enumerate}
        \item Recordings will undergo automated text–audio alignment and manual spot-checking.
        \item Only clear, intelligible recordings closely matching the target text will be retained.
    \end{enumerate}
\end{enumerate}

\subsection{Data protection rights consideration.}
\label{Appendix:dataprotection}
Contributors assign their data to the creators, making the transfer irrevocable to ensure dataset continuity. However, contributors retain moral rights. Moreover, as voice recordings can be considered personal information, participants can invoke privacy laws (e.g., South Africa’s POPIA laws) to request limited data usage or corrections via a formal process.

\end{document}